\pdfoutput=1

\documentclass[11pt]{article}

\usepackage[final]{acl}

\usepackage{times}
\usepackage{latexsym}
\usepackage{colortbl} 
\usepackage{xcolor}   

\usepackage[T1]{fontenc}

\usepackage[utf8]{inputenc}

\usepackage{microtype}

\usepackage{inconsolata}

\usepackage{graphicx}

\author{First Author \\
  Affiliation / Address line 1 \\
  Affiliation / Address line 2 \\
  Affiliation / Address line 3 \\
  \texttt{email@domain} \\\And
  Second Author \\
  Affiliation / Address line 1 \\
  Affiliation / Address line 2 \\
  Affiliation / Address line 3 \\
  \texttt{email@domain} \\}

\usepackage{microtype}
\usepackage{hyperref}
\usepackage{url}
\usepackage{booktabs}
\definecolor{darkblue}{rgb}{0, 0, 0.5}
\hypersetup{colorlinks=true, citecolor=darkblue, linkcolor=darkblue, urlcolor=darkblue}

\usepackage[T1]{fontenc}
\usepackage[utf8]{inputenc}
\usepackage{microtype}

\usepackage{comment}
\usepackage[many]{tcolorbox}
\usetikzlibrary{calc,shadows.blur}
\usepackage{filecontents}
\usepackage{enumerate}
\usepackage{framed}
\usepackage{subcaption}
\usepackage{wrapfig}
\usepackage{graphicx}
\usepackage{comment}
\usepackage{enumitem}
\usepackage{booktabs}
\usepackage{units}
\usepackage{multicol}
\usepackage[normalem]{ulem}
\usepackage{placeins}
\usepackage{wrapfig}
\usepackage{amssymb}
\usepackage{pifont}
\usepackage{multirow}
\usepackage{arydshln} 
\newtcbox{\inlinebox}[1][]{
 box align=base,
 nobeforeafter,
 colback=yellow!20,
 colframe=orange!50,
 size=small,
 left=0pt,
 right=0pt,
 boxsep=0.6pt,
 #1}

\usepackage{algorithm}
\usepackage{algorithmicx}
\usepackage[noend]{algpseudocode}
\usepackage{array}
\newcommand{\PreserveBackslash}[1]{\let\temp=\\#1\let\\=\temp}
\newcolumntype{C}[1]{>{\PreserveBackslash\centering}p{#1}}
\newcolumntype{R}[1]{>{\PreserveBackslash\raggedleft}p{#1}}
\newcolumntype{L}[1]{>{\PreserveBackslash\raggedright}p{#1}}

\usepackage{fontawesome}
\usepackage{tikz}

\newcommand{\todo}[1]{{\color{red} [TODO: {#1}]}}

\newcommand{\daniel}[1]{\textcolor{blue}{[DK: #1]}}
\newcommand{\datasetName}{\textsc{TurkingBench}}

\newcommand*\circled[1]{\textcircled{\fontsize{7pt}{0}\fontfamily{phv}\selectfont #1}}

\usepackage{color}
\usepackage{colortbl}
\definecolor{intnull}{RGB}{225, 245, 164}
\definecolor{intyellow}{RGB}{250, 244, 135}
\definecolor{intgray}{RGB}{233, 239, 240}
\definecolor{intred}{RGB}{245, 224, 201}
\definecolor{patterngreen}{RGB}{102, 212, 149}
\definecolor{semanticblue}{RGB}{152, 227, 226}

\definecolor{myblue}{rgb}{0.82, 0.94, 0.75}
\definecolor{mygreen}{rgb}{0.64, 0.76, 0.68}
\definecolor{myyellow}{rgb}{0.88, 0.54, 0.35}
\definecolor{mygreen}{rgb}{0.68, 0.9, 0.8}
\definecolor{mypink}{rgb}{0.2, 0.87, 0.2}
\def\arrvline{\hfil\kern\arraycolsep\vline\kern-\arraycolsep\hfilneg}

\def\mystrut(#1,#2){\vrule height #1pt depth #2pt width 0pt}   

\definecolor{purple}{rgb}{0.5,0,1}
\definecolor{dcyan}{rgb}{0.2,0.6,0.5}
\definecolor{light-gray}{gray}{0.95} 

\definecolor{darkgreen}{RGB}{0,140,0}
\definecolor{darkred}{RGB}{200,0,0}
\definecolor{lightgreen}{RGB}{197, 237, 208}
\definecolor{lightred}{RGB}{255,239,242}
\definecolor{lightyellow}{RGB}{255,240,160}
\definecolor{lightblue}{RGB}{225,241,255}
\definecolor{lightpurple}{RGB}{232,209,255}
\definecolor{lightgray}{RGB}{205,205,205}
\definecolor{indigo}{RGB}{13,165,240}

\definecolor{lightergray}{RGB}{230,230,230}
\definecolor{DarkRed}{RGB}{130,25,0}
\definecolor{DarkGreen}{RGB}{30,130,30}

\newcommand{\cmark}{\multirow{1}{*}{\textcolor{DarkGreen}{\ding{51}}}}
\newcommand{\xmark}{\multirow{1}{*}{\textcolor{red}{\ding{55}}}}

\newcommand{\LLaMA}{LLaMA2}
\newcommand{\GPTFour}{GPT4}

\newcommand{\TT}{\setlength{\fboxsep}{3pt}\colorbox{lightblue} T}
\newcommand{\VV}{\setlength{\fboxsep}{3pt}\colorbox{lightred} V}

\title{
\vspace*{-0.5in}
{{\small \hfill NAACL '25}\\
\vspace*{.25in}} 
\textsc{Tur[k]ingBench}: A Challenge Benchmark for Web Agents}

\usepackage{listings}
\usepackage{xcolor}

\lstdefinestyle{pythonstyle}{
    language=Python,
    basicstyle=\ttfamily\footnotesize,
    keywordstyle=\color{blue}\bfseries,
    stringstyle=\color{red},
    commentstyle=\color{green!60!black},
    showstringspaces=false,
    numbers=left,
    numberstyle=\tiny\color{gray},
    stepnumber=1,
    numbersep=10pt,
    frame=single,
    breaklines=true,
    backgroundcolor=\color{gray!10}
}

\newcommand{\mouseOne}{$^\text{\faMousePointer}$}
\newcommand{\handPointerZ}{$^\text{\mouseOne}$}

\newcommand{\handPointerC}{$^\text{\faICursor}$}
\newcommand{\handPointerA}{$^\text{\faHandPointerO}$}

\newcommand{\nameSpace}{\hspace{0.6cm}}
\newcommand{\nameSpaceB}{\hspace{0.6cm}}

\author{Kevin Xu\handPointerZ$^*$ \nameSpace Yeganeh Kordi\handPointerC \nameSpace Tanay Nayak\handPointerZ \nameSpace Adi Asija\handPointerZ \nameSpace \\ 
\textbf{Yizhong Wang\handPointerA \nameSpaceB Kate Sanders\handPointerZ \nameSpaceB  Adam Byerly\handPointerZ \nameSpaceB 
Jingyu Zhang\handPointerZ 
} \\ 
\textbf{Benjamin Van Durme\handPointerZ 
\nameSpaceB Daniel Khashabi\handPointerZ}\thanks{\;\;Correspondence to: \texttt{\{kxu39,danielk\}@jhu.edu} } 
\\
\handPointerZ Johns Hopkins University \nameSpaceB \handPointerC Brown University  \nameSpaceB \handPointerA University of Washington 
}

\begin{document}

\maketitle

\begin{abstract}


Can advanced multi-modal models effectively tackle complex web-based tasks? Such tasks are often found on crowdsourcing platforms, where crowdworkers engage in challenging micro-tasks within web-based environments.

Building on this idea, we present \emph{\datasetName}, a benchmark consisting of tasks presented as web pages with textual instructions and multi-modal contexts. Unlike previous approaches that rely on artificially synthesized web pages, our benchmark uses natural HTML pages originally designed for crowdsourcing workers to perform various annotation tasks. Each task's HTML instructions are instantiated with different values derived from crowdsourcing tasks, creating diverse instances. This benchmark includes 32.2K instances spread across 158 tasks.
To support the evaluation of \datasetName, we have developed a framework that links chatbot responses to actions on web pages (e.g., modifying a text box, or selecting a radio button). We assess the performance of cutting-edge private and open-source models, including language-only and vision-language models (such as GPT4 and InternVL), on this benchmark. Our results show that while these models outperform random chance, there is still significant room for improvement. We hope that this benchmark will drive progress in the evaluation and development of web  agents.\footnote{
Data and code: \url{https://turkingbench.github.io/}
}

\end{abstract}

\section{Introduction}

\newcommand{\specialcell}[2][c]{
\begin{tabular}[#1]{@{}c@{}}#2\end{tabular}
}

Significant progress in AI model development has been fueled by large pre-trained language models (LLMs)~\citep{radford2019language,dubey2024llama}. However, humans usually engage with visually rich environments, like web-based applications that combine language with visual elements such as images and tables. Notably, many people now spend more time online than in the physical world~\citep{perrin2019three}.



Crowdsourcing platforms are a key domain where human workers complete micro-tasks in exchange for rewards. Each day, hundreds of requesters publish tasks that need annotation by workers. These tasks are typically presented as web pages, incorporating stylistic features and structured information (e.g., tables, colors, font sizes) to communicate the tasks effectively. Inspired by humans' ability to solve various tasks through web interfaces~\citep{efrat2020turking}, and with the growing capabilities of multi-modal chatbots~\citep{openai2023gpt4, google2023gemini} that can generate and interact with external tools~\cite{chen2021evaluating,dubey2024llama,schick2023toolformer}, we are motivated to assess their performance on the same web-based crowdsourcing tasks that humans handle daily.



\begin{figure*}
    \centering    
    \includegraphics[scale=0.12,trim=2.1cm 81.9cm 120.0cm 8.5cm, clip=true]{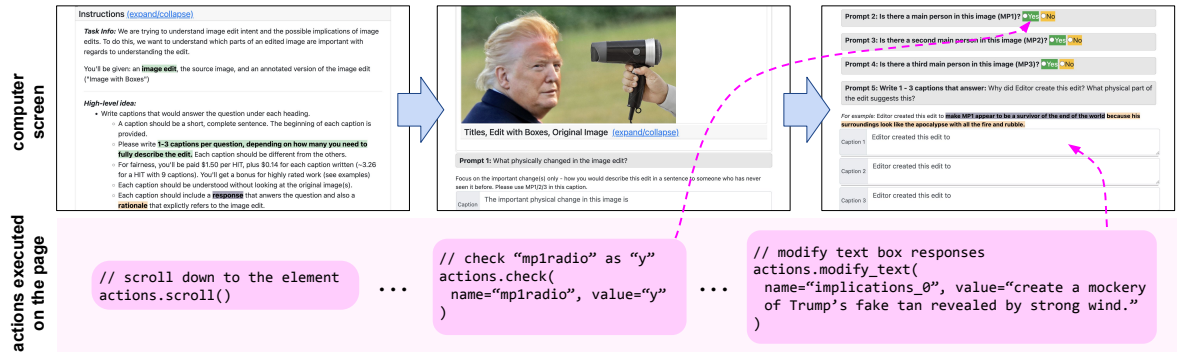} 
    \caption{
    An example of a web page (task) from \datasetName{}. Each page typically start with a few paragraphs of instructions and examples. Each task features a web page rich in diverse elements: tabular content organization, examples and target instances, color-coding for emphasis, bounding boxes around key instructions, multiple text boxes, images of people, and more. Naturally sourced from the wild for human users, these tasks encompass complex, interactive, and multi-modal reasoning for various web-based activities. Our benchmark motivates the development of web-based agents capable of processing such tasks and interactively filling in elements like radio buttons, check marks, and text boxes.
        Bottom row (in pink) shows the expected sequence of actions be executed  required  to  to solve the task (detailed in \S\ref{subsec:actions}).  
        More examples in Fig.\ref{fig:example}. 
    }
    \label{fig:actions:visualization}
\end{figure*}

We present \datasetName{}, a benchmark for web-grounded tasks. 
An example task 
can be seen in \autoref{fig:actions:visualization}. The information in these tasks is conveyed through a mix of stylistic elements (such as colors, font sizes, and shapes), structural features (like tables and paragraph headings), and multiple modalities (including text, images, and videos). Effectively interpreting these multi-modal signals is a complex challenge that demands a range of skills, particularly the ability to understand language within rich visual contexts.

The tasks in \datasetName{} are drawn from those published on the Amazon Mechanical Turk (AMT) platform, a well-known site where human workers complete micro-tasks for rewards. The benchmark comprises 158 web-grounded tasks (web pages), each averaging 16.8K tokens and containing 15.6 input fields that AI models must label or fill in. Each task can be instantiated with different input values (e.g., varying questions, paragraphs, images). In total, our benchmark includes 36.2K input fields.
There are several challenges in solving our proposed task:
(i) Multi-modality—tasks require models to ``see'' or ``read''.
(ii) Interaction—agents' actions must consider the context of previous decisions on the web pages.
(iii) Long context—web pages (HTML, screenshots, etc.) often are quite long. 

To facilitate interaction and evaluation of AI systems on these web-based tasks, we developed a Python-based middleware that connects AI models with web pages and also serves as the evaluation framework. This development, led by several students, has taken over a year of dedicated effort to ensure smooth functionality.


We use \datasetName{} to benchmark several notable unimodal and multi-modal models for interactive reasoning on web pages (\S\ref{subsec:results}). Among our evaluated models, GPT4 performs significantly better than random chance, either through visual actions (click and type) or textual commands that modify the HTML code of each page. Among the open-source models, Qwen2 achieves the highest performance.
In both settings, the top-performing models leverage the combination of text and visual modalities, highlighting the advantages of multi-modal approaches. Despite these positive gains, GPT4's performs well below the estimated upper bound of the dataset. We analyze model performance based on field type and task length, identifying challenges for future progress.


We chose crowdsourcing tasks for their diverse web interactions, including multimodal reasoning, complex form filling, and sequential dependencies—common in e-commerce, education, and administration. While rooted in this domain, our benchmark's challenges (e.g., diverse inputs, long-context reasoning, and multimodal interactions) apply broadly. Effective page understanding and manipulation remain key challenges for web agents.

Compared to existing web-based benchmarks, \datasetName{} fills a critical gap. A key feature of our benchmark is that both the web pages and instructions are naturally sourced from the wild, rather than being artificially constructed. The web pages in \datasetName{} were originally designed for human crowd workers, making them more authentic than recent benchmarks that use simplified, engineered web pages~\citep{liu2018reinforcement, furuta2023language}. Additionally, task instructions in \datasetName{} are embedded within the web pages, requiring a deep understanding of the content to solve them. This contrasts with most related benchmarks~\citep{deng2024mind2web, zhou2023webarena}, where instructions are provided as standalone sentences. Finally, because the tasks were initially created for complex annotation on crowdsourcing platforms (e.g., building NLP benchmarks), they inherently carry a high level of complexity. We believe this makes \datasetName{} a distinctive platform for evaluating both the comprehension and interaction capabilities of LLMs. 



\begin{table*}[ht]
    \centering
    \small
    \setlength{\tabcolsep}{0.5pt}
    \resizebox{\linewidth}{!}{
    \begin{tabular}{L{3.3cm}C{2.4cm}C{2.68cm}C{2.3cm}C{2.4cm}C{2.6cm}C{2.4cm}C{3.2cm}}
    \toprule
& MiniWoB++    \citep{liu2018reinforcement} & CompWoB   \citep{furuta2023language}  & RUSS \; \citep{xu2021grounding}  & WebShop \citep{yao2022webshop} & Mind2Web \citep{deng2024mind2web} & WebArena \citep{zhou2023webarena} & 
{\datasetName{} (this~work)} \\
 \cmidrule(lr){2-2} 
 \cmidrule(lr){3-3}
 \cmidrule(lr){4-4}
 \cmidrule(lr){5-5}
 \cmidrule(lr){6-6}
 \cmidrule(lr){7-7}
 \cmidrule(lr){8-8}
\rowcolor{lightergray!60}
\multirow{2}{*}{Inputs} & language commands & language commands & customer service queries & shopping queries & 
{language  commands} & English commands & instructions embedded in web pages \\
\specialcell{\hspace{-0.3cm}The input\\construction} & 
\specialcell{crowdsourced} & \specialcell{manual} & \specialcell{mined online} & \specialcell{crowdsourced} & \specialcell{crowdsourced} & 
\specialcell{author-written} &
\specialcell{collected\\from MTurk}
 \\
\rowcolor{lightergray!60}
Environments/domain & 
\specialcell{simplified\\ web page} & 
\specialcell{simplified but\\compositional}  & 
\specialcell{real-world\\ websites} & \specialcell{shopping\\web pages} & \specialcell{real-world\\websites} & \specialcell{variety of\\web pages} & \specialcell{crowdsourcing\\websites} 
\\
Natural inputs? 
& {\xmark} & \xmark & \xmark & \xmark & \xmark & \xmark & \cmark \\
\rowcolor{lightergray!60}
Realistic
interface?  & \xmark & \xmark & \cmark & \xmark & \cmark & \cmark & \cmark \\
Interaction w/ pages? & \cmark & \cmark & \xmark & \cmark & \xmark & \cmark & \cmark \\
\rowcolor{lightergray!60}
Functional correctness?  & \cmark & \cmark & \xmark & \cmark & \xmark & \cmark & \cmark \\
Inter-page navigation?  & \xmark & \cmark & \cmark & \cmark & \cmark & \cmark & \xmark \\
\bottomrule
\end{tabular}
}
\caption{Notable existing benchmarks for developing and evaluating web-based agents (\S\ref{sec:related}).
Existing benchmarks prioritize navigation between web pages but often use simplified or synthetic pages. Our work complements this by focusing on manipulating complex, naturally curated web pages, while excluding inter-page navigation.
}
\label{table:comparison}
\end{table*}

In summary, we introduce \datasetName{}, a benchmark for multimodal, interactive web-based tasks, along with an evaluation framework for web interaction. Our analysis of leading models highlights significant gaps, aiming to drive progress in assistive web agents.



\section{Related Work}
\label{sec:related}
Several benchmarks exist for evaluating web-navigation agents~\citep{lu2024weblinx,koh2024visualwebarena,he2024webvoyager,liu2024visualagentbench,pan2024webcanvas,cheng2023llf,zheng2024natural} (see \autoref{table:comparison}). These benchmarks offer reproducible model assessments but vary in their construction and coverage of real-world web navigation. Most focus on websites with limited complexity, such as \emph{pre-specified, synthesized}, or \emph{simplified} web pages~\citep{liu2018reinforcement,li2020mapping,yao2022webshop,furuta2023language}. In nearly all existing benchmarks, task definitions are created \emph{after} collecting the websites, typically through manual crowdsourcing. In contrast, \datasetName{} features more natural task instructions, originally intended for human users on crowdsourcing web pages. While this provides more natural task definitions, \datasetName{} is limited to tasks involving data annotation that require effective manipulation of complex web pages. Given the complementarity of these benchmarks, the research field is likely to benefit from all.

Unlike our focus on web pages, it is worth noting the benchmarks that utilize other environments for interactive problem solving. For instance, benchmarks that concentrate on mobile operating systems such as Android~\cite{li2020mapping,toyama2021androidenv,sun2022meta,burns2022dataset} or computer applications in operating system environments~\cite{trivedi2024appworld,xie2024osworld}.

\section{\datasetName: Benchmarking Web Agents via Multi-Modal
Turking Tasks}
\label{subsec:phase1}



We discuss the development of \datasetName{}. The primary purpose of this benchmark is to provide a standardized \emph{evaluation} for web-based agents. Additionally, a subset of this data can facilitate model design and development.

The overall development has undergone several rounds of meticulous engineering, led by multiple students, and has taken over a year of dedicated effort to ensure smooth functionality.

\subsection{Benchmark Schema}
\datasetName{} consists of a collection of tasks where 
each \emph{task} is a bundle of the following components: 
(A) A \textbf{web template} containing instructions for the task, input variables, and input fields for the outputs. 
 (B) \textbf{Input values} to be instantiated for the input variables.
 (C) Annotated \textbf{output labels} provided by crowd workers. 

\begin{figure}[ht]
    \centering
\includegraphics[scale=0.64,clip=true,trim=2.65cm 3.82cm 10.69cm 1.8cm]{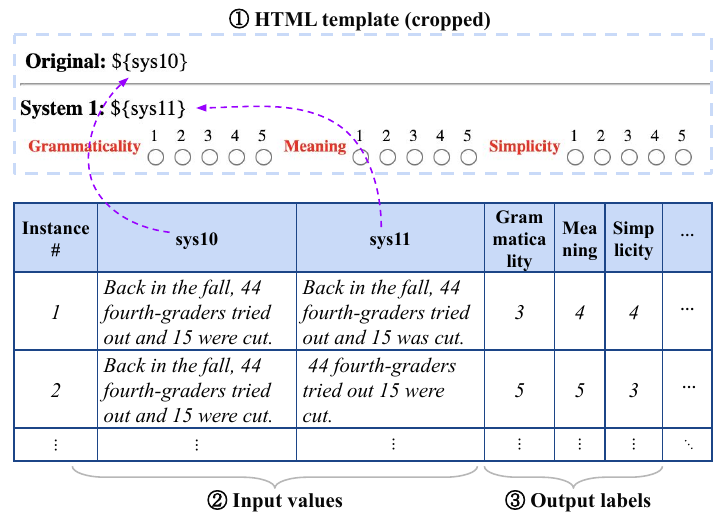} 
  \caption{
  An example showing the elements of our data: \circled{1} an HTML template with variables, \circled{2} input values from a CSV file populating the variables, and \circled{3} output labels used for evaluation obtained from crowdworkers.
  } 
  \label{fig:schema}
\end{figure}

An example is shown in \autoref{fig:schema}. As the example shows, the HTML template contains variables (e.g., \texttt{\$\{sys10\}}) that are instantiated with input values. 
Furthermore, the HTML template contains input fields for receiving input values from the web page users. 
For each field, we have values previously collected from crowd workers that we will use for evaluation. 

\subsection{Collecting the web-based tasks}

For a benchmark of tasks grounded in web pages to effectively measure progress and generalizability, it needs to be diverse and broadly covered. The majority of the tasks in \datasetName{} are sourced from prior crowdsourcing tasks conducted by the authors and their collaborators over the years. We also considered using tasks from the AMT sandbox,\footnote{\url{https://requestersandbox.mturk.com}} a testing environment for requesters to prototype their crowdsourcing templates. While this subset could have added more diversity to our data, it would have required additional annotation and cleanup, which we did not pursue.
During the selection process, we did not restrict ourselves to any specific task type. Our collected web pages encompass a variety of tasks, and this diversity is a strength. The wide range of tasks captures an array of problems that a generalist agent would encounter, mirroring real-world usability where tasks vary greatly.

We conducted light quality control to ensure the validity of the instructions and their annotations. Here are the steps we took:
(1) Several task instructions appeared underdefined (pilot experiments for crowdsourcing tasks), so we eliminated them.
(2) Certain tasks were missing data artifacts, such as images or videos. We eliminated these tasks if the missing media file was critical for solving the task.
(3) Some tasks did not render properly. A subset used older MTurk design conventions that are no longer supported. We manually revised or eliminated these tasks.
(4) Finally, we manually spot-checked the task annotations to ensure the quality was not noisy.

Ultimately, we put together a collection of 158 crowdsourcing tasks (examples in \autoref{fig:example}). 
These include but are not limited to language processing tasks (such as paraphrasing, validating factuality, entailment, sentiments, classification, dialogue quality, and rationale generation) and  tasks that involve processing images or videos. 



\begin{figure}[ht]
    \centering
    \includegraphics[scale=0.60,trim=0.8cm 0.3cm 0.0cm 0.0cm]{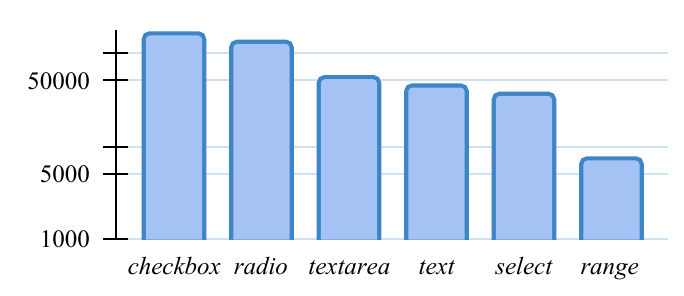}
     \caption{Distribution of input fields. }
  \label{fig:fields}
\end{figure}

\paragraph{Statistics.} The dataset contain a large sum of input-output instances (36.2K) across 158 tasks. 
\autoref{tab:dataset-summary} shows the overall statistics of \datasetName. 
Furthermore,  Figure~\ref{fig:fields} shows the distribution of the input fields represented in our benchmark. The most common field is ``checkbox,'' which is expected as it is commonly used when multiple options could be true. 
Here, we also distinguish between ``text'' input (a small, single-line box) and ``textarea'' input (a larger, multi-line box for descriptions and paragraphs) fields. 


\begin{table}[ht]
    \small
    \centering
    \begin{tabular}{lc}
       \toprule 
       Measure & Value \\ 
       \midrule
       \# of tasks & 158 \\
       \# of instances & 36.2K \\ 
       avg. \# of fields per task & 15.6 \\  
       avg. length (subwords) of the tasks & 16.8K \\  
       \bottomrule
    \end{tabular}
    \caption{Summary of dataset statistics. 
    }
    \label{tab:dataset-summary}
\end{table}

\subsection{Programmatic Interaction with Web }
\begin{figure}[ht]
  \centering
    \includegraphics[scale=0.22,trim=0.24cm 0.1cm 0.0cm 0.0cm, clip=true]{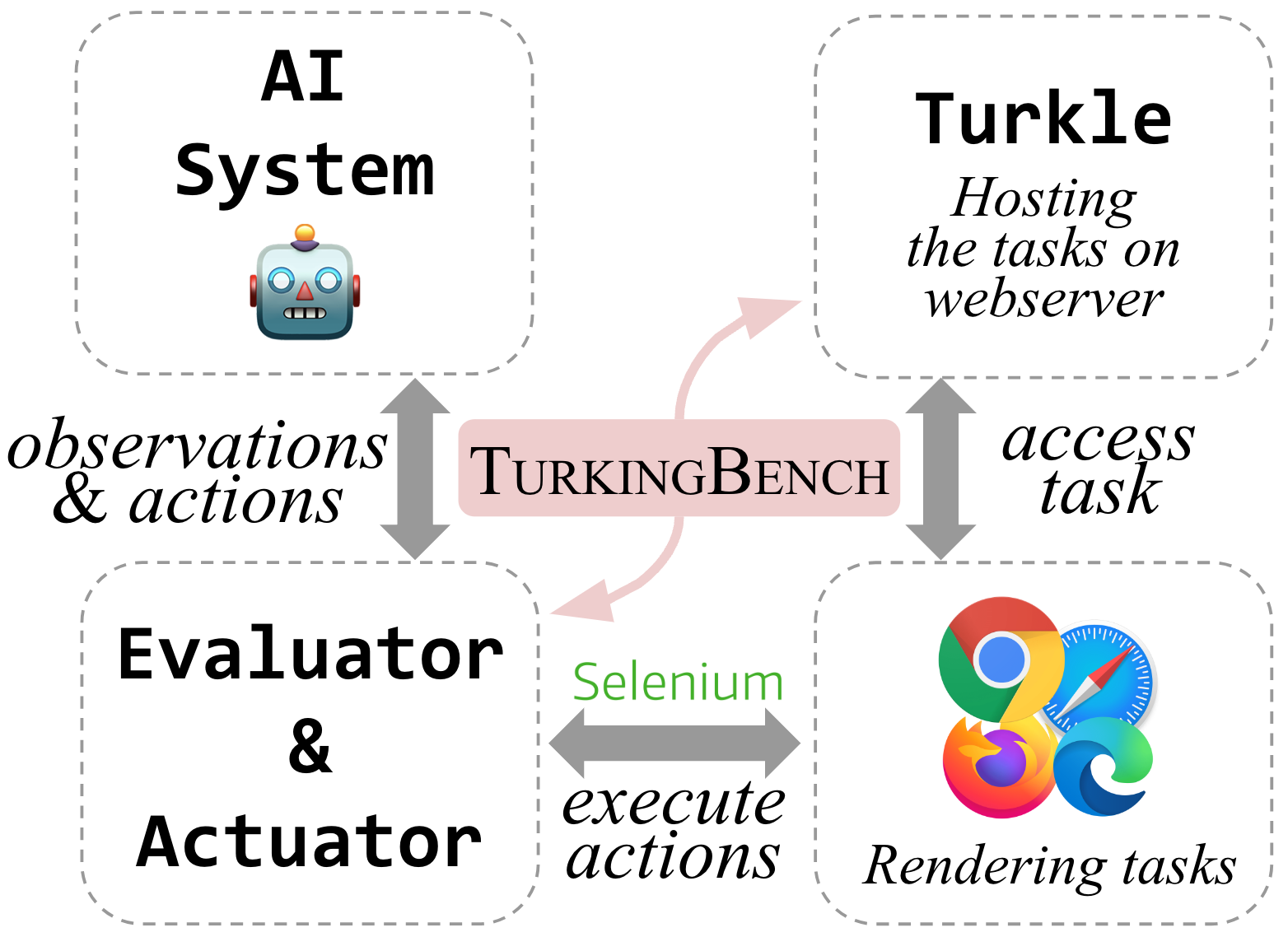} 
  \caption{
  The interface linking design segments: our web app serves data, which models access programmatically via our evaluation library.
    }
  \label{fig:framework}
\end{figure}

We developed a Python library to streamline the interaction of self-supervised models with our tasks and their evaluations (\autoref{fig:framework}). 
Our library contains various components. 
To serve our tasks, we use Turkle\footnote{\url{https://github.com/hltcoe/turkle}} which is an open-source replication of AMT.
The content of these tasks are then accessed through the web-browsers that are loaded by Selenium.\footnote{\label{footnote:selenium}\url{https://github.com/SeleniumHQ/selenium}}

Furthermore, we have developed a library of actions for accessing the pages' content and making changes to them. This is discussed in \S\ref{subsec:actions}. The starting point of interaction is our evaluation script that loops over the tasks and their instances. This is fleshed out in \S\ref{subsec:evaluation}

\subsubsection{A Library of Web Actions}
\label{subsec:actions}
To support model design, 
we have developed a library of ``actions'' that can execute various operations on a web page.
Our set of actions wrap around the API libraries provided by Selenium (c.f. \autoref{footnote:selenium}) and PyAutoGUI\footnote{\url{https://github.com/asweigart/pyautogui}} these are sophisticated libraries for web manipulation and  the amount of sophistication these libraries provide is significantly more than what is here. 
Therefore, we build our wrappers around these actions to build an action library with enough complexity to cover the dominant majority of tasks in our benchmark.

\begin{table}[ht]
    \centering
    \small
    \setlength{\tabcolsep}{2pt}
    \resizebox{\linewidth}{!}{
    \begin{tabular}{L{3.6cm}L{5.2cm}} 
        \toprule
        Actions (Modality) & \; \; \; Description\\
        \midrule
        \tt modify\_text (\TT) & modifies the text of input box  \\
        \tt modify\_checkbox (\TT) & modifies the selection of checkbox \\
        \tt modify\_radio (\TT) & modifies a radio button \\
        \tt modify\_select (\TT) & selects an item in a drop-down menu \\
        \tt modify\_range (\TT) & modifies a range input \\
        \tt get\_html (\TT) & fetches the HTML content of a page  \\
        \tt capture\_screen (\VV) & fetches the screenshot of a page \\
        \tt click (\VV) & clicks on a given coordinate  \\
        \tt type (\VV) & keyboard type at the selected  \\
        \tt scroll (N/A) & scrolls up or down  \\
        \tt maximize (N/A) & maximizes the web page \\
        \midrule
    \end{tabular}
    }
    \caption{The supported actions in our library.}
    \label{tab:actions}
\end{table}
The list of the actions included in our library is shown in \autoref{tab:actions}.
In terms of the modalities of information, a subset of these actions are either concerned with executing tasks in \colorbox{lightblue}{text (HTML)} modality or \colorbox{lightred}{visual} modality. 
There are fewer actions in the visual modality since visually many actions are combinations of clicking and typing. This is unlike the text modality where various element types require slightly different treatment.

In terms the nature of their roles, one can split these actions into three groups: 
(i) \emph{Modification} actions allow the models to modify each page's input elements. 
(ii) \emph{Navigation} actions allow the models to explore the page, for example, by scrolling up/down or waiting for all elements to be loaded. 
(iii) \emph{Sensing} actions allow models to retrieve the latest information about the web page, for example, by fetching its HTML code, taking screenshots of the active page, or around a given element. 

While our actions are designed to cover a broad set of web-based interactions, we acknowledge 
these are only a subset of what is needed for general-purpose navigation. Actions such as “drag-and-drop” are not included here since we did not have any tasks necessitating such actions. 


\paragraph{Web interaction as ``tool'' resolution.}
\label{subsec:tools}
The task here can be framed as iterative \emph{tool resolution} based on task context. A web agent must execute a sequence of executable actions from our library, guided by task instructions (see \autoref{fig:actions:visualization}). This aligns with tool-augmented LMs~\citep{schick2023toolformer,lu2024gear,mialon2023augmented,qin2023tool,gong2023arnold,lu2024toolsandbox}, where tools ground LM generations in an environment. Here, our library’s actions function as ``tools'' for achieving specific outcomes, with an LM aware of the tools, task context, and final goal.


\newcommand{\setOf}[1]{\left\lbrace #1 \right\rbrace}
\let\oldemptyset\emptyset
\let\emptyset\varnothing

\subsection{Evaluation Metrics}
\label{subsec:evaluation:metrics}

We devise an evaluation metric that is sensitive to the output type: 
    (i) \emph{Text fields:} For a given text response, we compute ROUGE metric~\citep{lin2004rouge} against each label and compute their maximum value. 
    (ii) \emph{Radio/select fields:} We compute an exact match of the predicted label  against the majority vote of gold labels. 
    (iii) \emph{Checkbox fields:} The results of the checkbox selection is a set. 
    Therefore, we quantify this as intersection over union metric between the two sets (set of predicted labels and the gold labels).  
    (iv) \emph{Range fields:} The range predictions provide real or integer values. We compute absolute ($\ell_1$) distance between the prediction and the labels, averaged across the labels. 
    This is normalized by the largest value among the gold labels to normalize it to $[0, 1]$ range. 

The overall score is an average of all the above responses. Given that the field distribution of each type does not vary from one model to another, averaging these different measures of quality is acceptable when comparing different models. 

\floatname{algorithm}{Pseudocode}

\algrenewcommand\algorithmicindent{0.99em}

\subsection{Evaluation Protocol}
\label{subsec:evaluation}
Our evaluation iterates through different choices of evaluation tasks (Pseudocode \autoref{alg:evaluation}). Specifically, the evaluation script supplies the model with the URL where the task is accessible.
Additionally, the model has access to our action library, enabling executable actions.

\begin{algorithm}[H]
  \caption{The evaluation protocol 
    \label{alg:evaluation}}
    \fontsize{9}{9}
    \selectfont
    \begin{algorithmic}
    \Require Action library: {\tt act} 
    \Require The evaluation tasks: {\tt tasks} 
    \Function{Evaluate}{{\tt tasks}, {\tt act}}, 
        \For{$t \gets$ {\tt tasks}}
            \State \textcolor{DarkGreen}{\# Solver receives each task, including its URL ({\tt t.URL})}
            \State \textcolor{DarkGreen}{\# and input fields ({\tt t.fields}). 
            It then executes actions}
            \State \textcolor{DarkGreen}{\# through the actions in {\tt act} (\autoref{tab:actions}).}
            \State \textsc{GenericSolver}{({\tt t}, {\tt act})} 
            \State \textcolor{DarkGreen}{\# Extract the new values of each field in {\tt t.fields} }
            \State \textcolor{DarkGreen}{\# Evaluate the predicted labels vs. gold labels (\S\ref{subsec:evaluation:metrics}). }
        \EndFor
    \EndFunction
    \end{algorithmic}
\end{algorithm}

Based on this evaluation protocol, it should be evident that our benchmark does not require navigation between web pages, as noted in the limitations section. However, our benchmark does require multiple rounds of interaction to solve a given task. This is because nearly all of our tasks involve more than one step (input field), each needing to be addressed in a different round of interaction.

To reduce the difficulty of the task for our models, we provide the list of field names on which we will evaluate model responses. While this arguably simplifies the task, our benchmark remains open for future analysis with open-ended interaction, without any additional guidance on the names of the target fields.

\begin{figure*}[ht]
    \centering
\includegraphics[scale=0.7,clip=false,trim=0.0cm 0.5cm 0cm 0cm]{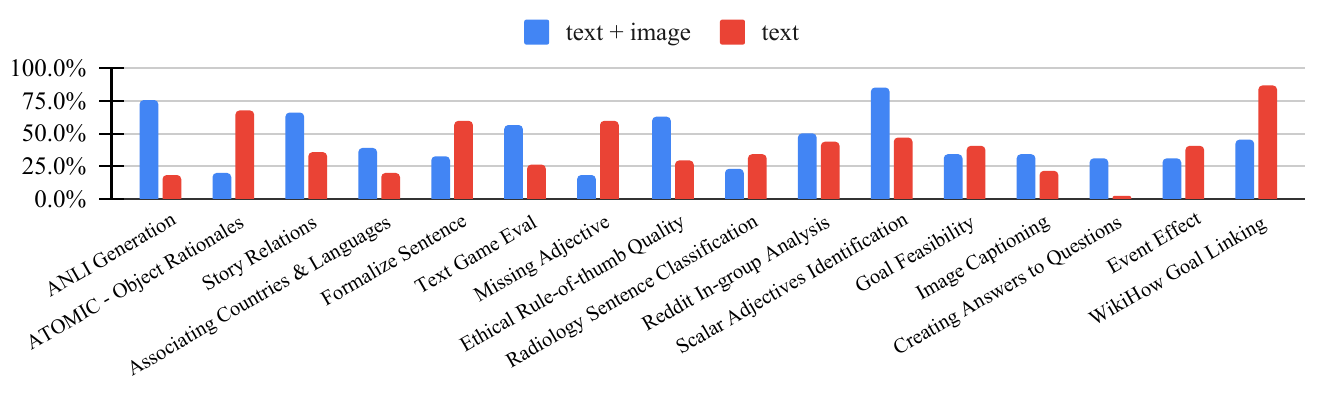}
  \caption{
    Performance of GPT4 (7 demonstrations) with two different input modalities across different tasks. 
  } 
  \label{fig:performance_per_tasks}
\end{figure*}

\paragraph{Task splits for measuring generalization to \emph{unseen} instructions.}
To accommodate for scenarios that may involve fine-tuning on supervised data, we create different task splits: 
(i) \emph{Train}: a set of 125 tasks that can be used for supervised model development. 
(ii) \emph{Test}: a set of 16 tasks that  be used for evaluating model quality (names shown in the caption of Fig.~\ref{fig:performance_per_tasks}). 
(iii) \emph{Test}$_{\text{challenge}}$: a set of 17 tasks that are strictly more difficult than the rest of the tasks, because they require more complex combination of actions or other tasks not covered in \autoref{tab:actions}.
In this version of the work, we do not provide any experiments on \emph{Test}$_{\text{challenge}}$ since they require more sophisticated engineering that is beyond the scope of this work and should be addressed in future work.

\paragraph{Screen Size Considerations.}  
When benchmarking multimodal models, the {\tt capture\_screen} action retrieves visual information from the screen. A potential concern is the impact of screen size on benchmarking consistency. To address this, all evaluations using {\tt capture\_screen} run in headless mode, where the browser operates without a monitor and uses predefined, consistent virtual screen dimensions. This eliminates hardware-specific influences, ensuring robust and reproducible results unaffected by screen size variability.







\section{Evaluating Models in Solving Web-based Tasks in \datasetName}
\label{subsec:models}
We benchmark various models with different architectures on \datasetName. Our goal is to provide reasonable baselines for our proposed benchmark, so we avoid specialized models that rely on specific assumptions or supervision of our data.

\begin{table*}[ht]
    \centering
    \scriptsize
    \setlength{\tabcolsep}{4pt} 
    \small
    \resizebox{0.99\linewidth}{!}{
    \begin{tabular}{ccccccccc}
       \cmidrule[1pt]{1-9}
      &  \specialcell{Model} &  \specialcell{\# of Parameters\\ (model size)} & Modalities & \specialcell{Text\\encoding} & \specialcell{\# of \\demos} & \specialcell{Input len\\(subwords)} &  \specialcell{Maximum Input \\Sequence Length} & Score (\%)    \\ 
       \cmidrule[0.5pt]{1-9}
        \cellcolor{intgray} & \cellcolor{intgray} Do-nothing & \cellcolor{intgray} --  &  \cellcolor{intgray} -- & \cellcolor{intgray} -- &  \cellcolor{intgray} -- & 
        \cellcolor{intgray} -- & 
        \cellcolor{intgray} -- & 
        \cellcolor{intgray} 7.8 \\
        \cmidrule[0.3pt]{1-9}
       \multirow{13}{*}{\rotatebox{90}{{open-source models}}} & \multirow{2}{*}{Llama3.1-Instruct} & 8B &  \TT & relevant & 3 & 
       1k & 128k & 23.2  \\
       & & 8B & \TT & relevant & 7 & 5k & 128k & 25.0 \\
       \noalign{\vskip 0.5ex}
       \cdashline{2-9}
       \noalign{\vskip 0.5ex}
       & \multirow{2}{*}{Llama3.2-Instruct} & 3B &  \TT & relevant & 3 & 
       1k & 128k &  20.8  \\
       & & 3B & \TT & relevant & 7 & 5k &  128k &  20.3 \\
       \noalign{\vskip 0.5ex}
       \cdashline{2-9}
       \noalign{\vskip 0.5ex}
       & \multirow{2}{*}{Llama3.2-Vision} & 11B &  \TT + \VV & relevant & 3 & 
       1.2k & 128k &  9.28  \\
       & & 11B & \TT + \VV & relevant & 7 & 6k &  128k &  17.8 \\
       \noalign{\vskip 0.5ex}
       \cdashline{2-9}
       \noalign{\vskip 0.5ex}
       & \multirow{2}{*}{LLaVA-VL-Vicuna} & 7B &  \TT + \VV & relevant & 3 & 
       1.6k & 4096 &  22.7  \\
       & & 13B & \TT + \VV & relevant & 3 & 1.6k & 4096 &   19.8 \\
       \noalign{\vskip 0.5ex}
       \cdashline{2-9}
       \noalign{\vskip 0.5ex}

       & \multirow{2}{*}{InternVL2} & 40B &  \TT + \VV & relevant & 3 & 
       1.6k & 8192 &  31.0  \\
       && 40B & \TT + \VV & relevant & 7 & 8k & 8192 & 
  26.1 \\
        \noalign{\vskip 0.5ex}
       \cdashline{2-9}
       \noalign{\vskip 0.5ex}
       & \multirow{2}{*}{InternVL2} & 76B & \TT + \VV & relevant & 3 & 1.6k & 8192 & 
  30.4 \\
        && 76B & \TT + \VV & relevant & 7 & 8k & 8192 & 
  29.6 \\
   \noalign{\vskip 0.5ex}
       \cdashline{2-9}
       \noalign{\vskip 0.5ex}
      & \multirow{1}{*}{Qwen2} & 72B &  \TT + \VV & relevant & 7 & 
           6k & 128k &  \textbf{34.1}  \\
        \cmidrule[0.6pt]{1-9}
       \multirow{15}{*}{\rotatebox{90}{{closed-source models}}} & \multirow{1}{*}{Claude2.1} & Unknown &  \TT & full &   7 & 
       82k & 200k & 22.6  \\
       \noalign{\vskip 0.5ex}
       \cdashline{2-9}
       \noalign{\vskip 0.5ex}
       & \multirow{3}{*}{GPT4} & \multirow{3}{*}{Unknown} & \TT & relevant & 1  & 1.2k & 128k &  19.2\\
       &   & & \TT & relevant & 3  & 4k & 128k &  18.4\\
       &   &  & \TT & relevant & 7  &  5k & 128k &  21.3 \\
       \noalign{\vskip 0.5ex}
       \cdashline{2-9}
       \noalign{\vskip 0.5ex}
       & \multirow{3}{*}{GPT4} & \multirow{3}{*}{Unknown} & \TT & full & 1  & 21k & 128k &  18.7\\
       &   &  & \TT & full & 3  &  28k & 128k &  35.7 \\
       &   &  & \TT & full & 7  &  82k & 128k &  39.3 \\
        \noalign{\vskip 0.5ex}
        \cdashline{2-9}
        \noalign{\vskip 0.5ex}
        & \multirow{3}{*}{GPT4o} & \multirow{3}{*}{Unknown} &  \TT + \VV & relevant & 0 & 
       900 & 128k &  32.4\\
        &  &   & \TT + \VV & relevant & 3 & 
       1.6k & 128k &  30.3 \\
        &  &   & \TT + \VV & relevant & 7 & 
       8k & 128k &  30.19 \\
       \noalign{\vskip 0.5ex}
       \cdashline{2-9}
       \noalign{\vskip 0.5ex}
       & \multirow{3}{*}{GPT4-V} & \multirow{3}{*}{Unknown} &  \TT + \VV & full & 1 & 
       22k & 128k &  19.4\\
       &  &   & \TT + \VV & full & 3 & 
       30k & 128k &  41.1  \\
        &  &   & \TT + \VV & full & 7 & 
       86k & 128k &  \textbf{41.7}  \\
       \cmidrule[0.3pt]{1-9}
       \cellcolor{intgray} & \cellcolor{intgray}  Oracle & \cellcolor{intgray} -- & \cellcolor{intgray} -- & \cellcolor{intgray} -- &
       \cellcolor{intgray} -- &
       \cellcolor{intgray} -- &
       \cellcolor{intgray} -- &
       \cellcolor{intgray} 100.0 \\
       \cmidrule[0.8pt]{1-9}
    \end{tabular}
    }
    \vspace{-0.1cm}
    \caption{
        Comparison of language (\TT) and vision-language (\VV) model performance on \datasetName. We evaluate two approaches to encoding HTML documents: "full" includes the entire HTML content, while "relevant" includes only a few neighboring lines of HTML adjacent to the target input field. Due to context window limits, most open-source models could not accommodate more than 3 demos, all using "relevant" text encoding. Among the combinations explored, \GPTFour{} outperforms most open-source models but remains far from our ceiling performance.
    }
    \label{tab:models_performance}
\end{table*}

\paragraph{Models.} As discussed earlier, our models need to consume information on a mix of information modalities. A model can, therefore, consume the instructions as text, image, or a combination of both. 
We experiment with state-of-the-art proprietary models like GPT4 and Claude-2.1.\footnote{GPT4/GPT4-V/Claude accessed via APIs in January 2024; GPT4o accessed in August 2024.} 
For GPT4 models, we experiment with both the text models and the vision-language variant that is trained in a joint of visio-textual information. 
We compare the performance of these models with the latest open-source vision language models like LLaVA-1.6 \citep{liu2023llava}, InternVL2 \citep{chen2023internvl} and text-only models like Llama-3.1-Instruct~\cite{dubey2024llama} and Qwen2~\cite{qwen2}.
Our evaluation focused on major model families (GPT4, LLaVA1.6, etc.), as our goal was setting baselines with widely-recognized models. 

We note that, besides these general-purpose models, there are more specialized text-based models for web exploration, either through pre-training on text data~\citep{aghajanyan2021htlm,aghajanyan2022cm3,understandinghtmlIzzedin2022}, specialized models for analyzing web-based components~\citep{huang2022layoutlmv3,tao2022knowing,ebrahimi2023test,chen2022xdoc}, or visual perception of the web~\citep{dosovitskiy2021image,rust2023language,pix2struct2022lee,kil2024dual}. We leave such explorations, which are likely to yield better results, to future work.

\paragraph{Encoding the tasks for evaluation.}
When processing the HTML content of the tasks, we consider two variants: 
(1) ``full'' indicates all the HTML content of the entire page.
(2) ``relevant'' indicates a few neighboring lines of HTML adjacent to (above and under) the input field the model is currently solving (further details in \S\ref{app:relevant}.)
To reduce the cost of evaluation, we use 20 instances per each of the 20  evaluation tasks. This adds up to $(20\times20=)400$ web pages evaluated with a total of roughly $6k$ input fields evaluated.

\paragraph{An oracle for ceiling performance.}
We implement an oracle baseline that mimics the gold labels for each input (see the logic in \autoref{alg:oracle-pseodocode}). \footnote{See the a \href{https://youtu.be/eT2OhU_Nodg}{screencast of the oracle's execution.} \label{footnote:screencast}}
This oracle agent ensures the functional correctness of our evaluation. Our end-to-end evaluation process (model output $\rightarrow$ execution $\rightarrow$ lookup answers from web pages $\rightarrow$ evaluation) is significantly more complex than a typical NLP benchmark, and any of these steps can fail. Achieving 100\% accuracy with the oracle (ensuring functional correctness) took the lead student several months of effort.
The oracle baseline essentially replicates the action sequences of crowdworkers for each HTML input element in the task web pages, executing appropriate actions similar to those shown in \autoref{fig:actions:visualization}.






We note that the oracle baseline is limited to tasks that do not require complex annotations (such as drag-and-drop) included in \emph{Test}$_{\text{challenge}}$ (discussed in our evaluation split~\S\ref{subsec:evaluation:metrics}). A future use case of this oracle baseline can be to obtain granular action sequences for supervising models.

\paragraph{A ``do-nothing'' model as a lower-bound.}
We evaluate a trivial baseline that 
performs no action (no actions (hence, ``do-nothing'').
As we see in the results, this baseline scores more than zero  because on some tasks doing nothing is the right action (e.g., making grammatical corrections to a given text that happens to be grammatical). 

\subsection{Empirical Results}
\label{subsec:results}
We present the results of evaluating our main models in \autoref{tab:models_performance}. We experimented with models of varying parameter sizes. For each row, we indicate whether the model input contains text-only (\TT) or text-vision (\TT+\VV). Additionally, the table shows the size of input prompts measured in GPT-2 subwords~\cite{radford2019language}.
Wherever possible, we evaluate the models with 7 in-context demonstrations of tasks and desired actions. The open-source vision-language models (\TT+\VV) had much shorter context windows, so we evaluated them using the ``relevant'' portion of the HTML code for the task or in-context demonstrations.



\paragraph{Despite the remarkable performance of generalist models, they remain far from our ceiling performance.}
The best performance $41.7\%$ is obtained by GPT4 vision-language model (\TT+\VV) with access to full text of each task. 
This is in-line with other recent observations~\cite{zhenggpt}.
This configuration also happens to have a extremely large prompt length (86k subwords) and it shows the remarkable ability of this model to exploit long-range dependencies. 
We note that the gains of the vision model (\TT+\VV) over text-only model (\TT) is minimal ($41.7$ vs. $39.5$).

\paragraph{Open-source models rivaling GPT4.}
Llama3.1-Instruct (8B params) notably outperforms GPT4 (text-only), achieving a score of 25.0\% compared to GPT4’s 21.3\% with 7 demonstrations when ``relevant'' bits of the input HTML are supplied. This result is particularly impressive given that Llama3.1-Instruct operates with significantly fewer parameters (8B) than GPT4's rumored size. The comparison highlights Llama3.1-Instruct’s ability to efficiently leverage the provided context, potentially making it better suited for certain tasks despite its smaller size.
Additionally, Qwen2 (72B) demonstrates strong performance, coming very close to GPT4 when evaluated with the ``relevant'' HTML. Qwen2’s score of 34.1\% is not too far off from GPT4-V’s 41.7\%, which underscores the potential and promise of open-source models over  proprietary models. 

\subsection{Analysis: Test vs Vision Models}

For one of our top configurations in \autoref{tab:models_performance} (GPT-4, 7 demonstrations with ``full'' HTML encoding), we present a breakdown of model performance across tasks in \autoref{fig:performance_per_tasks}.  On 9 out of 16 evaluation tasks, the two models exhibit notable performance differences.  
While it's intuitive to assume tasks with richer interfaces benefit more from visual input, empirical results do not clearly pinpoint which task design aspects drive these differences.  
Overall, we conclude that \textbf{text-only and vision-language models have complementary capabilities}.

\subsection{Analysis: Effect of the Number of Demos }

\begin{figure}[ht]
    \centering
\includegraphics[scale=0.75,clip=false,trim=0.4cm 0.3cm 0cm 0cm]{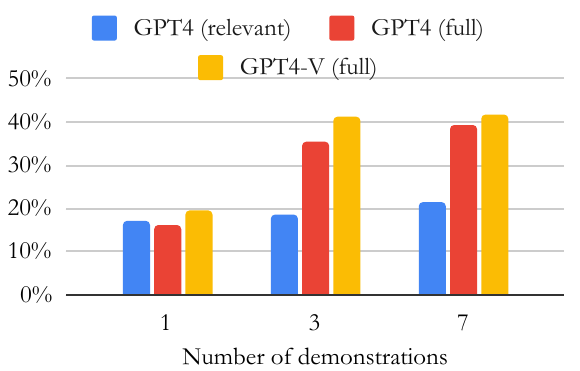} 
  \caption{
    Performance with varying number of demos. 
  } 
  \label{fig:perf:vs:shots}
\end{figure}

As mentioned earlier, we use few-shot prompting of models in order to steer them their predictions. 
Here we study the effect of the number of demonstrations in model performance. 
As the results are shown in \autoref{fig:perf:vs:shots}, \textbf{the gains of in-context demonstrations quickly plateaus} when the number of demonstrations just above 3 demonstrations. 

\subsection{Analysis: Performance Per Field Types}

We provide extended details on performances per input field type in \autoref{tab:model_performance}.  
Note that we distinguish between "text" and "textarea" fields because they are defined with different HTML tags (\texttt{<input type="text">} vs. \texttt{<textarea>}). The former is typically used for short inputs, while the latter is for longer, multi-sentence outputs. 
There is no consistent patter here. Most models tend to have similar weaknesses and strengths, which is not surprising as they’re generally trained on similar-ish [pre-training] data. 
A notable finding was the surprisingly lower performance of GPT4 compared to other models on ``text'' input fields, which we could not explain. However, since the data is skewed toward checkboxes (\autoref{fig:fields}), GPT4 performs better in the aggregate evaluations. 
This suggests the potential for hybrid models by combining those that excel at specific input field types.

\begin{table}[h]
    \setlength{\tabcolsep}{3pt} 
    \small
    \centering
    \resizebox{1.05\linewidth}{!}{
    \begin{tabular}{lC{0.8cm}C{1cm}clccccc}
        \toprule
        Model & \# of params & \# of demos & Modality & all & checkbox & radio & text & textarea \\
        \midrule
        Llama-3.2  & 3 B  & 3  & \TT     & 20.8  & 58.7  & 25.8  & 19.1  & 0.3  \\
        Llama-3.2  & 3 B  & 7  & \TT     & 20.3  & 56.8  & 22.9  & 22.0  & 2.9  \\
        Gemma-2    & 7 B  & 3  & \TT     & 20.4  & 54.1  & 28.7  & 15.1  & 2.0  \\
        InternVL   & 40 B & 3  & \TT + \VV & 30.7  & 59.0  & 85.0  & 46.0  & 12.0 \\
        InternVL   & 40 B & 7  & \TT + \VV & 26.1  & 85.0  & 28.2  & 48.1  & 7.5  \\
        InternVL   & 76 B & 3  & \TT + \VV & 30.4  & 63.7  & 29.7  & 48.4  & 17.1 \\
        InternVL   & 76 B & 7  & \TT + \VV & 29.6  & 56.6  & 30.5  & 46.4  & 12.3 \\
        GPT4-VL    & -    & 3  & \TT + \VV & 30.3  & 73.3  & 34.2  & 16.4  & 15.5 \\
        \bottomrule
    \end{tabular}
    }
    \caption{Performance comparison of models across different modalities and input types.}
    \label{tab:model_performance}
\end{table}










\subsection{Error Analysis}
We conducted human annotations to better understand the results in \autoref{tab:models_performance}. This analysis uses predictions from GPT4o (7 demos and ``full'' encoding) as it is one of the highest-performing models. One of the authors reviewed one instance from each of the 16 tasks, totaling 144 responses to the input fields.


\paragraph{Evaluating GPT4 responses.}
Our annotator directly evaluated 134 (out of 144) responses that we successfully parsed by our evaluation metric (i.e., no parsing error). 
The annotator agreed with GPT4o's responses in 60\% (80 out of 134) of the cases, reaffirming that our benchmark remains challenging for the models.

Our analysis revealed several recurring issues. One common problem occurred in binary classification tasks, such as determining whether a word is an adjective with a similar meaning to a given word. For example, GPT4o incorrectly classified ``discernible'' as describing ``modernity,'' despite the lack of synonymy.

Another significant discrepancy was observed in tasks requiring GPT4o to generate both correct and incorrect answers. Instead of providing actual incorrect answers, GPT4o sometimes returned placeholders like \texttt{//incorrect options//}, failing to meet the task’s requirements. In some cases, GPT4o also made syntax errors. However, the evaluation focused on the content of the responses rather than their syntactical correctness, so syntax errors did not affect the assessment of answer accuracy. 
In \autoref{tab:examples} we highlight examples of model predictions for a given UI.

\section{Conclusion and Future Work}
We introduced \datasetName{} to facilitate research on web-based agents. Our benchmark focuses on tasks defined within the context of web pages, such as those commonly found on crowdsourcing platforms. It includes a comprehensive Python-based framework that supports both evaluation and model development. We hope this benchmark will drive further advancements in the development of web-based assistant agents.


Future work should explore modeling improvements for better web agents. For instance, a RAG-style approach could semantically chunk web pages into meaningful segments, a non-trivial task. Another avenue could involve using these agents as CoPilots for human annotation on crowdsourcing platforms, helping workers identify potential mistakes. We consider these extensions somewhat orthogonal to the primary focus of this work and hope future research will address them.

\section*{Limitations}
We discuss several limitations: 
\paragraph{Intra page navigation.} Our benchmark does not require navigation between web pages. While inter-page navigation is important, it is not the only challenge. 
Effective understanding and manipulation of each page, which is our focus, remains a significant challenge for web agents. 
Our benchmark still requires navigation within each page. 
This is because almost all of our tasks have more than one step (input field), each needed to be solved in a different round of interaction.
As shown in our experiments, the benchmark remains challenging for the models. We accept this trade-off to obtain more natural tasks compared to most existing benchmarks.

\paragraph{Simplified evaluation.} This is not an inherent limitation of our benchmark but a simplifying assumption in our evaluation. For simplicity, our evaluation setup provides some guidance by specifying the names of the fields to be modified. Future work should explore variants of our experiments where such hints are not provided to the model.

\paragraph{Scope.}
Our benchmark's distribution is biased to crowd-sourcing tasks and hence it is not a holistic measure of web-based agents.
That being said, 
\textit{all} benchmarks carry their own biases. 
In the context of web-based tasks, all the existing benchmarks (WebShop, Mind2Web, WebArena, etc.) have their own set of assumptions and restrictions about the scope of web-based agents. We view all these efforts to be  complementary, each quantifying a unique aspect of intelligent behavior on the web.

\section*{Ethical Considerations}
We recognize concerns that this work could lead to technologies replacing crowd workers, who are vital to AI development. This
concern is not unique to our work extends to all AI. Our results show that state-of-the-art models are still far from fully automating crowdsourcing tasks, even with simplified evaluation. Therefore, we hope our work enables benevolent use-cases of AI such as enhancing the quality and productivity of crowd workers rather than replacing them.

\section*{Acknowledgements}
This project is partly supported by ONR grant (N00014-24-1-2089) and a generous gift from Amazon.
The authors would like to thank 
Chris Callison-Burch, 
Mark Dredze, 
Karthik Narasimhan, 
Nikhil Sharma, 
Owen Bianchi, 
and Elizabeth Salesky
for their constructive discussions. 
GPU machines for conducting experiments are provided by the ARCH (Rockfish) cluster (\url{https://www.arch.jhu.edu}).

\bibliography{ref}

\clearpage

\appendix 

\begin{center}
{\Large \textbf{Supplemental Material}}
\end{center}

\section{Additional Related Work}

Here we discuss other related work that did not fit in the main text. 

\paragraph{Multi-modal reasoning tasks.}
\datasetName{} is also related to efforts in multi-modal interactive environments~\citep{gur2018learning,ku2020room,li2020mapping,li2022mug,li2022spotlight,sun2022meta,li2021vut,bai2021uibert}. However, these often feature simple instructions that are only a few sentences long, unlike our more extensive instructions embedded within web pages.




\paragraph{Web-based agents.}
The concept of intelligent automated assistant agents collaborating with humans to complete tasks has been around for some time~\citep{allen2007plow} and can be seen as an extension of early work on semantic parsing~\citep{das2010probabilistic,clarke2010driving,bordes2012joint,understandinghtmlIzzedin2022}. Recent literature has explored various forms of supervision, including behavior cloning of actions~\citep{gur2023real}, reinforcement learning~\citep{liu2018reinforcement,nakano2021webgpt,humphreys2022data,liu2023agentbench}, and in-context learning (ICL)~\citep{kim2023language,tao2023webwise,sridhar2023hierarchical}. Our work focuses on introducing a new benchmark, providing ICL baselines, and leaving the exploration of more sophisticated models for future research.

\section{Pseudocode for the oracle baseline}

\algnewcommand\algorithmicswitch{\textbf{switch}}
\algnewcommand\algorithmiccase{\textbf{case}}
\algnewcommand\algorithmicassert{\texttt{assert}}
\algnewcommand\Assert[1]{\State \algorithmicassert(#1)}%
\algdef{SE}[SWITCH]{Switch}{EndSwitch}[1]{\algorithmicswitch\ #1\ \algorithmicdo}{\algorithmicend\ \algorithmicswitch}%
\algdef{SE}[CASE]{Case}{EndCase}[1]{\algorithmiccase\ #1}{\algorithmicend\ \algorithmiccase}%
\algtext*{EndSwitch}%
\algtext*{EndCase}%

\algrenewcommand\algorithmicindent{0.99em}

\algnewcommand{\parState}[1]{\State%
    \parbox[t]{\dimexpr\linewidth-\algmargin}{\strut #1\strut}}

\begin{algorithm}[H]
  \caption{The oracle baseline 
    \label{alg:oracle-pseodocode}}
    \fontsize{9}{9}\selectfont
    \begin{algorithmic}
    \Require Action library: {\tt act} 
    \Require The target fields to be modified: {\tt fields} 
    \Require The gold labels for each field: {\tt labels} 
    \Function{OracleSolver}{{\tt fields}, {\tt labels}}
        \For{$f \gets$ {\tt fields}}
            \State {\tt act.wait\_till\_loaded($f$)}  
            \State {\tt act.scroll\_to($f$)} 
            \State $\ell \gets {\tt labels}(f)$ 
              \Switch{$f.$type}
                \Case{{\tt text}: \;\;\;\; Execute {\tt act.modify\_text($f$,$\ell$)}} 
                \EndCase
                \Case{{\tt radio}: \;\; Execute {\tt act.modify\_radio($f$,$\ell$)}} 
                \EndCase
                \Case{{\tt select}:  Execute {\tt act.modify\_select($f$,$\ell$)}} 
                \EndCase
                \Case{{\tt range}: \; \;Execute {\tt act.modify\_range($f$,$\ell$)}} 
                \EndCase
              \EndSwitch
        \EndFor
    \EndFunction
    \end{algorithmic}
\end{algorithm}

\section{Extracting the ``relevant'' HTML code for each field}
\label{app:relevant}
We detail the procedure for retrieving "relevant" neighboring HTML, as discussed in \S\ref{subsec:models} under "Encoding the tasks for evaluation."
The “relevant” HTML encoding is implemented by extracting and returning a subset of HTML adjacent to (above and under) the specific input field on a webpage.Specifically: (1) We retrieve the full HTML content of the target elements' grandparents elements. This potentially contains a lot of content.
(2.) We split this code into html elements (split based on occurrence of {\tt ">"}) (3) We select 15 elements before and 30 elements below the target element. 
The original function is accessible in our public \href{https://github.com/JHU-CLSP/turking-bench/blob/f1a385459517c42c84ad9552d8820accd45e0c64/src/evaluation_class.py#L625-L648}{code base}.

\onecolumn

\begin{table}[ht]
    \centering
    \small
    \begin{tabular}{L{0.97\textwidth}}
        \toprule
        
        \textbf{Category:} 
        Both automatic and human evaluations classified the action as \textit{incorrect} (true negative).
        \\ 
        \includegraphics[scale=0.5]{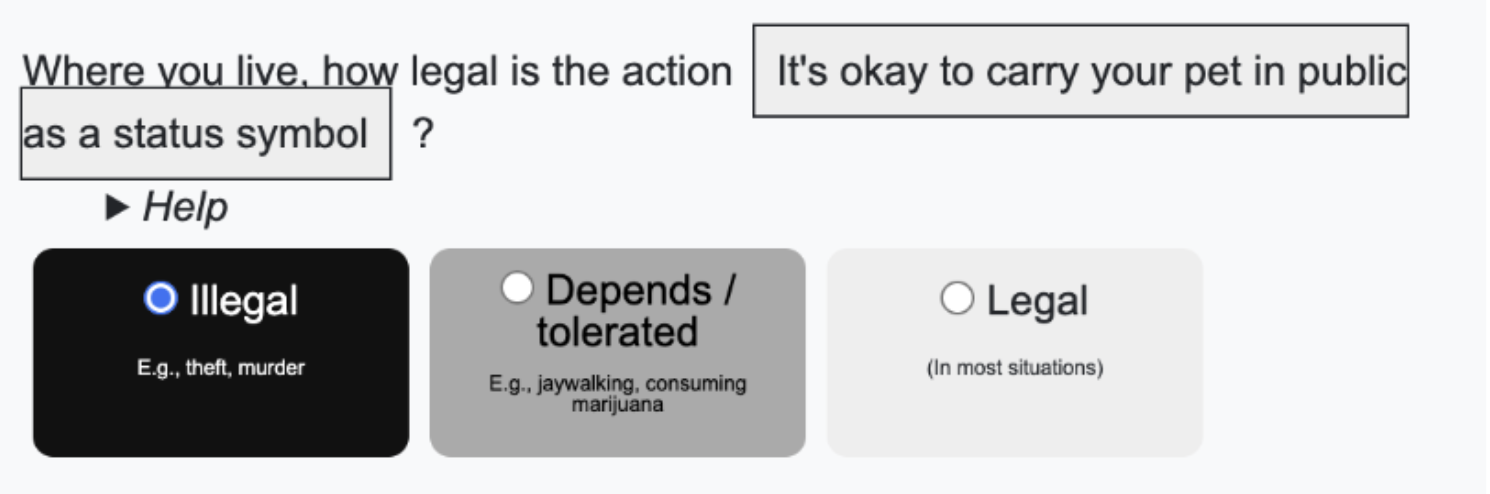} \\
        \textbf{Explanation:}
        It is perfectly legal to do carry your pet in public, even as a status symbol.\\ 

        \midrule

        \textbf{Category:} 
        Both automatic and human evaluations classified the action as \textit{incorrect} (true negative).
        \\ 
        \includegraphics[scale=0.5]{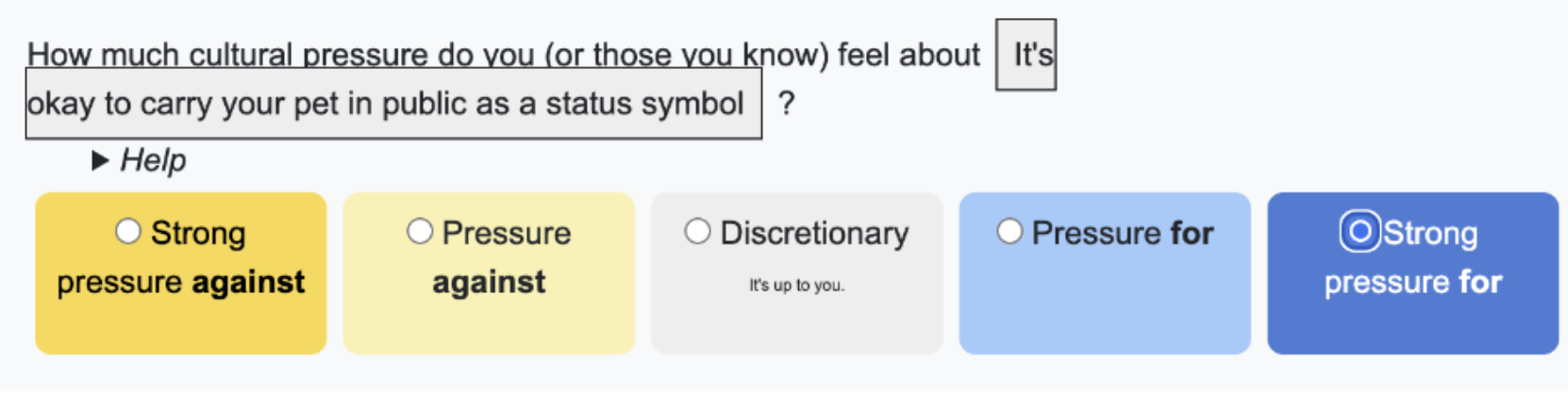} \\
        \textbf{Explanation:}
        Our society does not "strongly pressure" people to carry pets in public.\\ 

        \midrule

        \textbf{Category:} 
        Automatic evaluation rated the action as \textit{incorrect} but human evaluations rated it \textit{correct} (false negative).
        \\ 
        \includegraphics[scale=0.5]{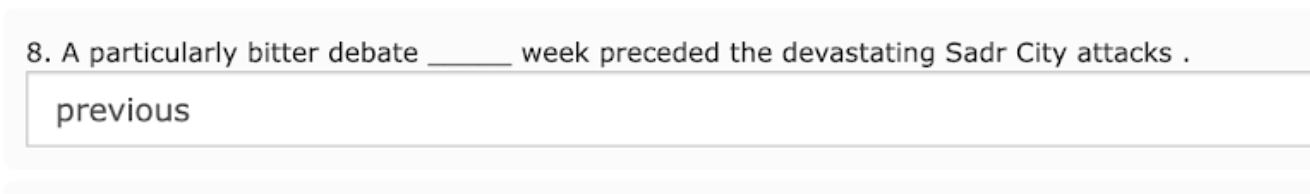} \\
        \textbf{Explanation:}
        The automatic evaluation has rated it \textit{incorrect} since `previous' was not found in the list of gold answers: 
        ['filled', 'this', 'last']. 
        \\ 
                \midrule

        \textbf{Category:} 
        Automatic evaluation rated the action as \textit{incorrect} but human evaluations rated it \textit{correct} (false negative).
        \\ 
        \includegraphics[scale=0.5]{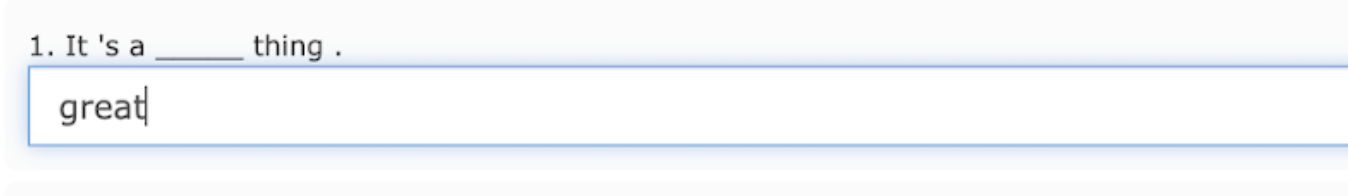} \\
        \textbf{Explanation:}
        The automatic evaluation has rated it incorrect since `great' was not found in the list of gold answers: 
        ['good', 'guy', 'girl', 'Good', 'big']. 
        \\ 
        \midrule

        \textbf{Category:} 
        Automatic evaluation and human evaluation both rated the action as \textit{correct}
        (true positive).
        \\ 
        \includegraphics[scale=0.5]{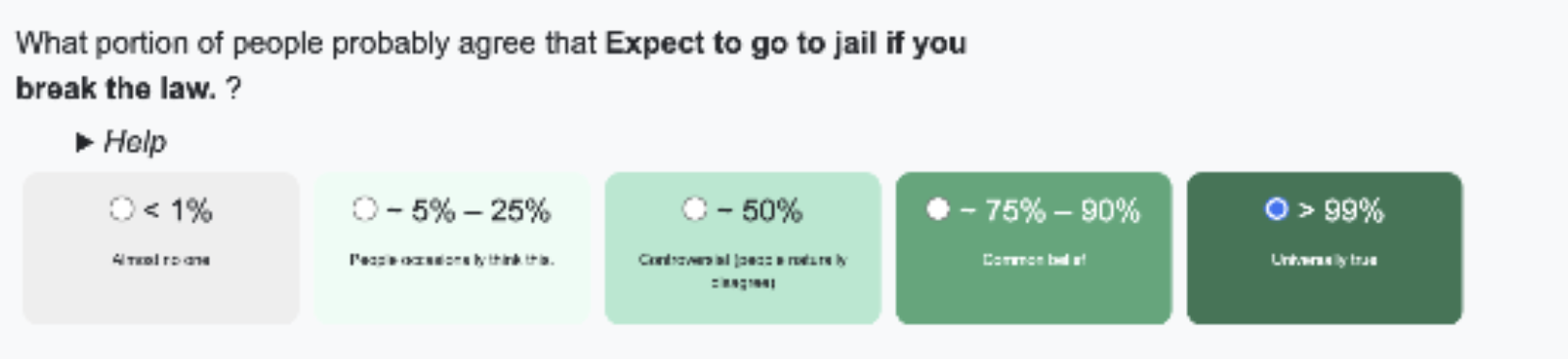} \\
        \textbf{Explanation:} -- 
        \\ 
        \midrule
                \textbf{Category:} 
        Automatic evaluation and human evaluation both rated the action as \textit{correct}
        (true positive).
        \\ 
        \includegraphics[scale=0.35]{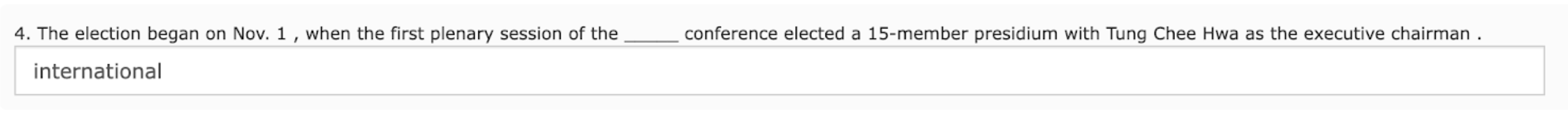} \\
        \textbf{Explanation:} -- 
        \\ 

                \bottomrule
    \end{tabular}
\end{table}

\begin{table}[ht]
    \centering
    \small
    \begin{tabular}{L{0.97\textwidth}}
        \toprule

        \textbf{Category:} 
        Automatic evaluation rated the action as \textit{correct} even though human evaluations rated it \textit{incorrect} (false positive).
        \\ 
        \includegraphics[width=0.99\textwidth]{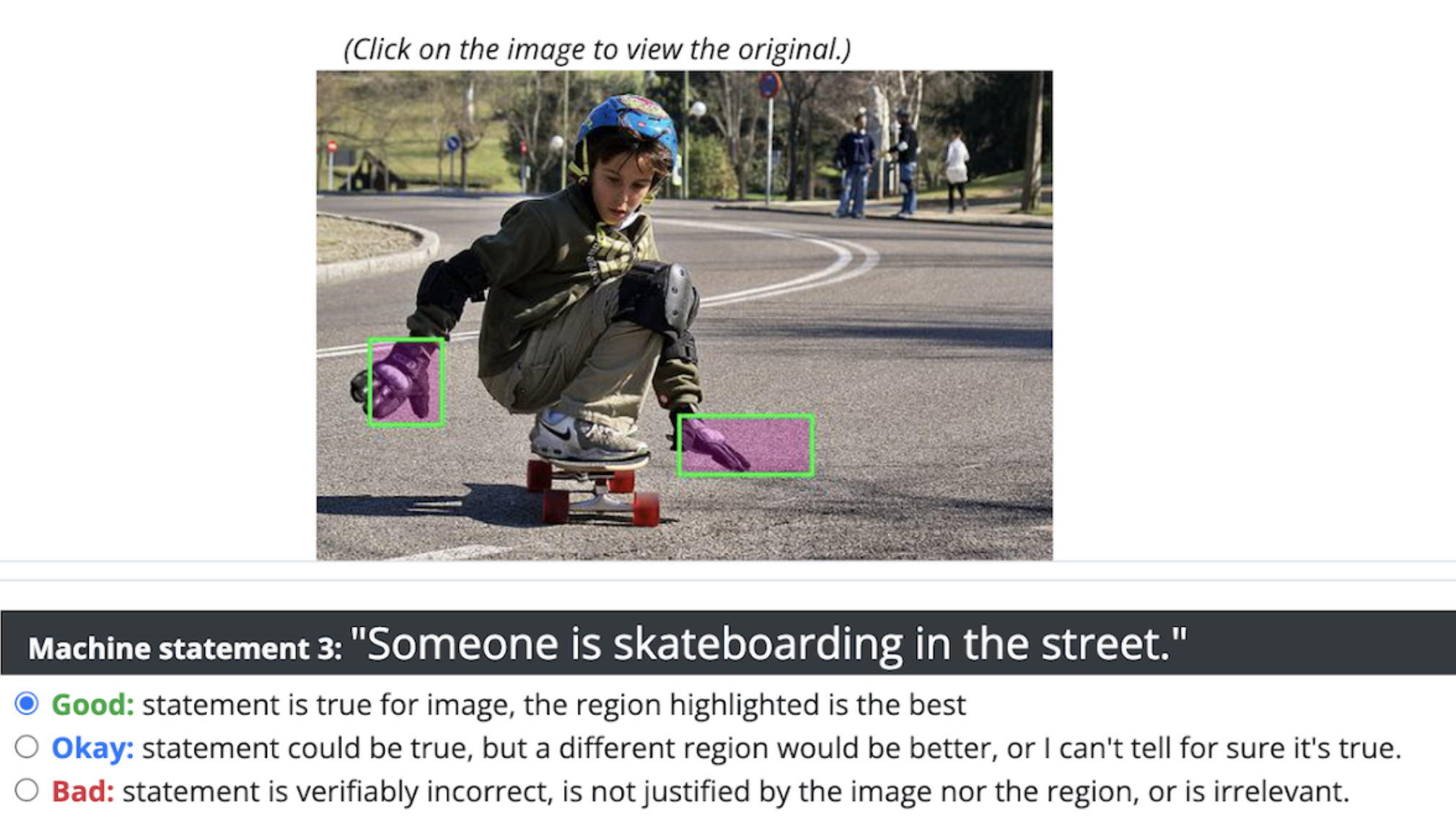} \\
        \textbf{Explanation:}
        The statement is true, but the highlighted region is irrelevant. 
        \\ 
                \midrule

                \textbf{Category:} 
        The model response does not adhere to the expected syntax. 
        \\ 
        \includegraphics[width=0.97\textwidth]{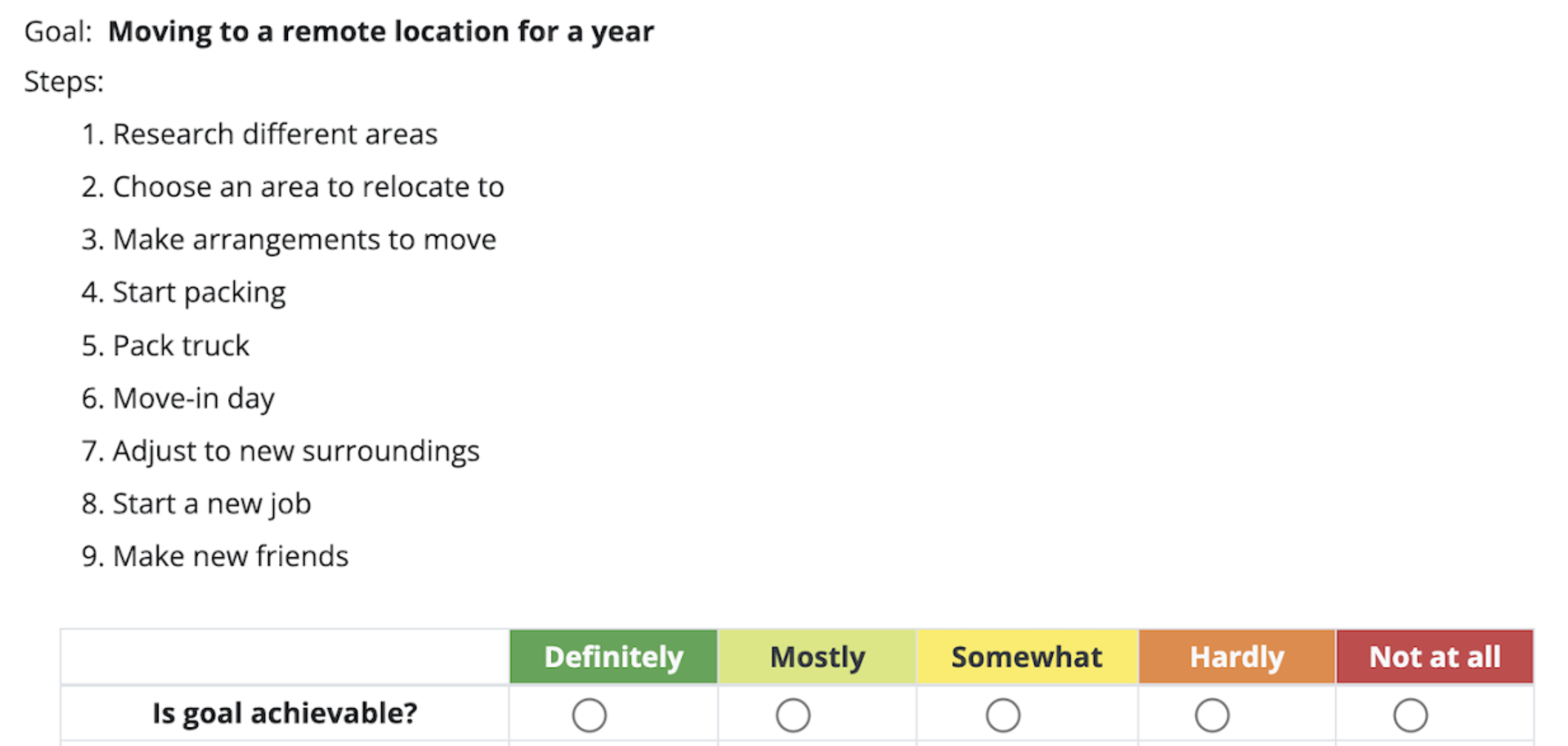} \\
        \textbf{Explanation:}
        The execution here failed since instead of selecting one of the radio values from -2 to 2, the model responded with a blank string.
        
        \\ 

        \bottomrule
    \end{tabular}

    \caption{Examples predictions of GPT4-v which is the best performing model in \autoref{tab:models_performance}. For each example, we also highlight the outcome of automatic vs human evaluation.   }
    \label{tab:examples}
    
\end{table}

\begin{figure*}[ht]
    \centering
    \resizebox{0.97\linewidth}{!}{
    \includegraphics[scale=0.213, trim=7.11cm 53.2cm 172.7cm 6.15cm, clip=false]{figures/turk-instructions-examples.pdf}
    }
    \vspace{-0.1cm}
    \caption{
Examples of the web pages for several tasks included in \datasetName{} are shown. These pages typically start with a few paragraphs of instructions and examples. Each task features a web page rich in diverse elements: tabular content organization, examples and target instances, color-coding for emphasis, bounding boxes around key instructions, multiple text boxes, images of people, and more. Naturally sourced from the wild for human users, these tasks encompass complex, interactive, and multi-modal reasoning for various web-based activities. Our benchmark motivates the development of web-based agents capable of processing such tasks and interactively filling in elements like radio buttons, check marks, and text boxes.
    }
    \label{fig:example}
\end{figure*}

\end{document}